\PassOptionsToPackage{table}{xcolor}
\documentclass{article} 
\usepackage{iclr2027_conference,times}

\usepackage{amsmath,amsfonts,bm}

\def\eqref#1{equation~\ref{#1}}

\def\1{\bm{1}}

\DeclareMathAlphabet{\mathsfit}{\encodingdefault}{\sfdefault}{m}{sl}
\SetMathAlphabet{\mathsfit}{bold}{\encodingdefault}{\sfdefault}{bx}{n}

\renewcommand{\eqref}[1]{(\ref{#1})}

\usepackage[table]{xcolor}
\usepackage{colortbl}
\usepackage{graphicx}
\usepackage{booktabs}
\usepackage{tabularx}
\usepackage{multirow}
\usepackage{amsmath}
\usepackage{amssymb}
\usepackage{amsthm}
\usepackage{bm}
\usepackage{algorithm}
\usepackage{algpseudocode}
\usepackage{enumitem}
\usepackage{caption}
\usepackage{subcaption}
\usepackage{wrapfig}
\usepackage{placeins}
\usepackage{needspace}
\usepackage[hidelinks]{hyperref}
\usepackage{url}
\usepackage{marvosym} 

\graphicspath{{figures/}}

\newcommand{\ours}{QMap\-Anything}
\definecolor{oursgray}{gray}{0.92}
\newcommand{\oursrow}{\rowcolor{oursgray}}
\newcolumntype{Y}{>{\centering\arraybackslash}X}
\newcommand{\calD}{\mathcal{D}}
\newtheorem{proposition}{Proposition}
\newtheorem{lemma}{Lemma}

\title{Rotation Is Token-Blind: A Structural Criterion for Post-Training Quantization of Feed-Forward Geometry Transformers}

\author{%
\begin{minipage}{0.97\textwidth}
\centering\normalfont
Huanfei He\textsuperscript{1,2,*},
Bohao Zhang\textsuperscript{2,*},
Jiaxin Zhang\textsuperscript{1},
Wei Sui\textsuperscript{2,\ensuremath{\dagger}},\\[3pt]
John See\textsuperscript{4},
Zeyd Boukhers\textsuperscript{3}, and
Cong Yang\textsuperscript{1,\Letter}\\[8pt]
{\small
\textsuperscript{1}School of Future Science and Engineering, Soochow University,
Suzhou, China\\[2pt]
\textsuperscript{2}D-Robotics, Beijing, China\\[2pt]
\textsuperscript{3}Fraunhofer Institute for Applied Information Technology,
Sankt Augustin, Germany\\[2pt]
\textsuperscript{4}School of Mathematical and Computing Sciences,
Heriot-Watt University Malaysia,\\ Putrajaya, Malaysia\\[5pt]
\texttt{hfhe@stu.suda.edu.cn, wei.sui@d-robotics.cc, cong.yang@suda.edu.cn}}
\end{minipage}%
}

\iclrfinalcopy
\begin{document}

\maketitle
\lhead{Preprint}
\begingroup
\renewcommand{\thefootnote}{\fnsymbol{footnote}}
\footnotetext[1]{Equal contribution.}
\footnotetext[2]{Project leader.}
\renewcommand{\thefootnote}{\Letter}
\footnotetext[3]{Corresponding author.}
\endgroup

\begin{figure}[!ht]
\centering
\includegraphics[width=0.86\linewidth]{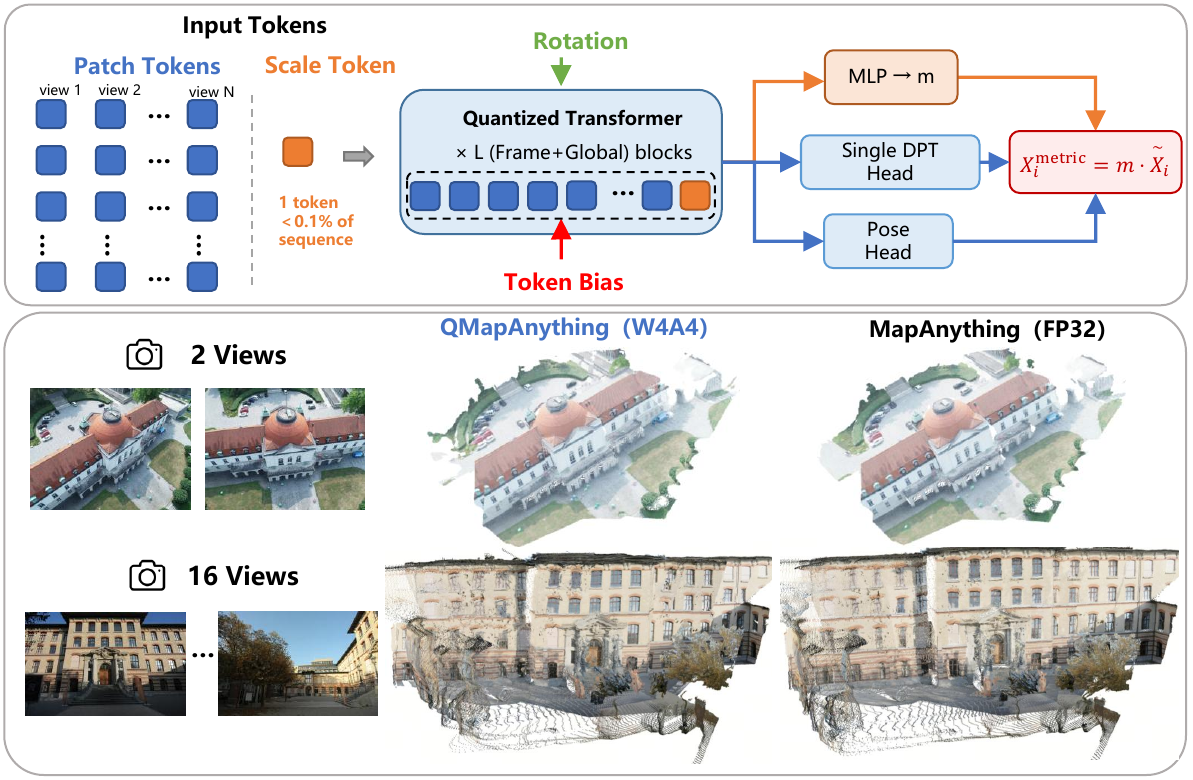}
\vspace{-4pt}
\caption{\ours{} overview. MapAnything's scale token occupies ${<}0.1\%$ of the sequence yet
shares one activation grid with $Nn$ patch tokens, from $N$ images with $n$ patches each. We pair a rotation matrix with a row-selective
affine correction. Bottom: W4A4 reconstructions at 2 and 16 views remain visually close to the
corresponding FP32 results.}
\label{fig:teaser}
\vspace{-8pt}
\end{figure}

\begin{abstract}
Feed-forward 3D models use Transformers to predict camera poses, depth, and structure in one pass,
avoiding iterative multi-view optimization. Long sequences and deep attention incur high
compute and memory costs, motivating Post-Training Quantization (PTQ) for deployment.
Existing PTQ methods mainly suppress channel-wise outliers by smoothing or orthogonal rotation. Geometry
Transformers introduce a different challenge: heterogeneous token types share one activation
quantization grid. A few special tokens may exert high leverage over global outputs and be updated
by only a subset of blocks. Token-blind transforms, which apply the same transformation to every token, reduce overall quantization error but cannot
remove differentiated degradation across token types.
We propose \ours{}, a structure-aware PTQ framework. We cast floating-point-preserving rotation
and smoothing as a token-blind reparameterization group, and prove that they yield common-mode gains while leaving relative token SNR unchanged. We introduce a statistic $\Gamma$ that combines a special token's quantization
disadvantage and output leverage to predict whether differentiated bias correction is needed
beyond rotation.
Experiments on geometric benchmarks show that W4A8 closely matches full-precision performance. Under W8A8, peak memory falls by
$2.2$--$3.1\times$ versus FP32, with up to $3.1\times$ speedup on real hardware. These results provide
a principled and efficient approach to deploying feed-forward 3D reconstruction models in
resource-constrained edges.
\end{abstract}

\section{Introduction}
\label{sec:intro}

Classical multi-view reconstruction combines geometric priors with iterative
optimization~\citep{orbslam2,colmap_sfm,colmap_mvs}. Learning-based systems first moved dense
matching and point-map prediction into neural networks~\citep{dust3r,mast3r}, and recent
feed-forward Transformers now recover camera pose, depth, and 3D structure jointly in a single
forward pass~\citep{vggt,mapanything}. This paradigm offers strong generalization without
per-scene optimization, but its long token sequences and deep attention stacks impose substantial
memory and computation costs, making post-training quantization (PTQ) attractive for deployment.

Recent PTQ methods suppress outliers through scaling, smoothing, or orthogonal
rotation~\citep{smoothquant,quarot,quipsharp,spinquant}. Related recipes have been extended
to visual geometry Transformers~\citep{quantvgt,qvggt,fgq}. These methods act along the
channel axis, whereas feed-forward geometry models can be asymmetric along the token axis. We
identify four properties that characterize this mismatch: \textbf{(S1)} heterogeneous token types
share one activation grid. \textbf{(S2)} A single token has multiplicative, output-wide leverage. \textbf{(S3)} Channel energy is heavy-tailed. \textbf{(S4)} The special token
is updated by only a subset of blocks. Rotation addresses S3, but its token-blind action
cannot remove the differentiated error created by S1--S2. Consequently, metric outputs may fail
even when scale-aligned geometry remains plausible, and a recipe successful on an up-to-scale
model need not transfer to a metric one.

We present \ours{}, a structure-aware PTQ framework that uses these properties to select mechanisms.
We first characterize all floating-point-preserving rotation and smoothing transformations as a
token-blind reparameterization group. Orthogonal transformations provide a common-mode signal-to-noise ratio (SNR) gain but
leave relative token SNR unchanged. More general transformations face a conditioning ceiling and a
potential weight-side quantization trade-off. We then introduce $\Gamma$, a screening statistic that combines a
token's grid disadvantage with its output leverage. It requires only one forward and
backward pass per scene and screens for the need for token-selective treatment. A residual statistic measured
during calibration determines affine correctability. Guided by this analysis, \ours{} pairs a
rotation matrix for S3 with standard bias correction and, where warranted,
a row-selective affine correction for S1--S2, inserted only where S4 makes it necessary. MapAnything
provides the primary stress test, while Visual Geometry Grounded Transformer (VGGT) and Depth Anything 3 (DA3) provide complementary transfer cases. VGGT has high gradient leverage
on its special token without a grid disadvantage, whereas DA3 has a grid disadvantage without effective
output leverage.
Our contributions are summarized as follows:
\begin{enumerate}[topsep=1pt, partopsep=0pt, itemsep=1.5pt, parsep=0pt]
\item We identify four structural properties (S1--S4) that
explain when channel-oriented PTQ transfers to a geometry Transformer and when it does not,
separating metric predictors from up-to-scale ones.
\item We prove that floating-point-preserving rotation and smoothing form a token-blind
group. They act identically on every token, so they have common-mode strength but no
row-indexed freedom. \ours{} keeps a rotation inside this group and adds a row-selective
affine correction outside it, inserted only where S4 indicates.
\item We introduce a screening statistic for that need and a residual statistic to test affine
correctability. The former distinguishes MapAnything from VGGT and DA3, and \ours{} attains
near-lossless low-bit accuracy with $3.1\times$ lower peak memory.
\end{enumerate}
\vspace{-6pt}

\section{Related Work}
\label{sec:related}
\vspace{-4pt}

\paragraph{Learning-based 3D reconstruction.}
Classical pipelines separate structure-from-motion and multi-view stereo (MVS)~\citep{orbslam2,colmap_sfm,colmap_mvs}.
Learned MVS~\citep{mvsnet} and cross-view pretraining~\citep{croco} move geometry into networks.
DUSt3R and MASt3R predict pairwise geometry~\citep{dust3r,mast3r}, VGGT jointly predicts pose and dense
geometry~\citep{vggt}, and MapAnything factors metric geometry into an up-to-scale representation
and one global scale~\citep{mapanything}. Beyond their differing prediction targets, these models
also differ in whether a special token is routed through every block and whether it multiplies
rather than merely summarizes the output. Those differences drive our analysis.

\paragraph{Quantization, viewed as reparameterization.}
Second-order compensation~\citep{gptq} is one PTQ line.
Rounding and reconstruction~\citep{adaround,brecq,hubara2021smallcalib,qdrop}, Transformer
scaling~\citep{zeroquant,llmint8,smoothquant,awq,omniquant,rptq,outliersuppressionplus,affinequant},
rotation or structured codebooks~\citep{quip,quipsharp,quarot,spinquant,atom,squeezellm,aqlm},
and Vision Transformer (ViT) calibration with nonlinear operators
~\citep{liu2021ptqvit,yuan2022ptq4vit,lin2022fqvit,ding2022apqvit,li2023repqvit,liu2023noisyquant,
wu2024adalog,yang2024dopqvit,wu2025fimaq,wu2025aphqvit} are others.
For geometry Transformers,
QuantVGGT combines rotation, smoothing, and fine-grained activations~\citep{quantvgt}. QVGGT uses
mixed precision, token filtering, and principal component analysis (PCA) compensation~\citep{qvggt}. FGQ instead uses diagonal
Fisher information for affine calibration~\citep{fgq}. Diagonal~\citep{smoothquant,awq,affinequant}
and orthogonal~\citep{quipsharp,quarot,spinquant,quantvgt} reparameterizations reduce common-mode
error but provide no criterion for when a token-blind recipe suffices. We add a pre-quantization criterion
$\Gamma$ that keeps in-group rotation and inserts a special-row term only when needed, so the same
rule transfers. Standard bias correction~\citep{banner2019,nagel2019dfq} is token-uniform and
misses that residual.


\begin{figure*}[t]
\centering
\includegraphics[width=\textwidth]{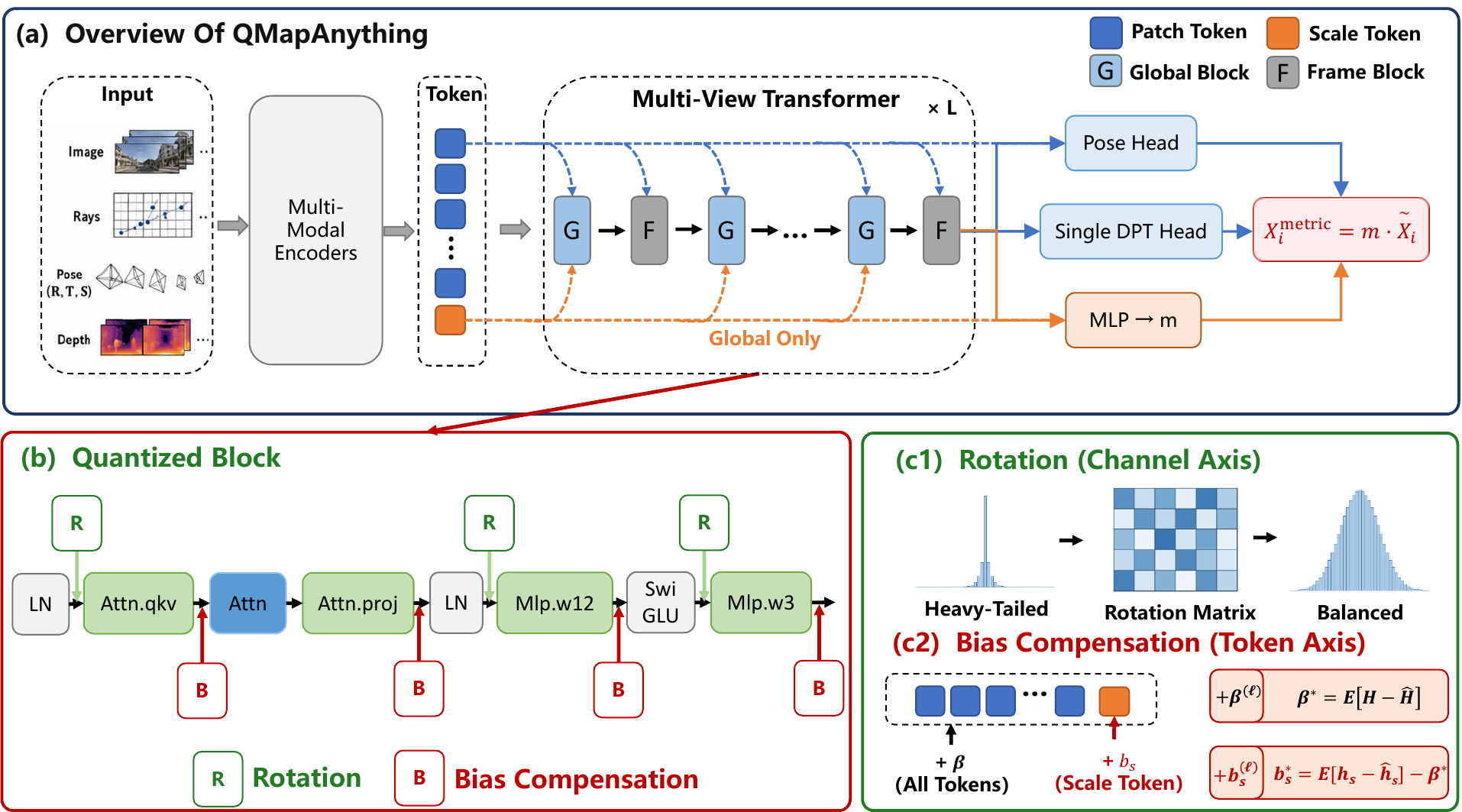}
\vspace{-8pt}
\caption{Overview of \ours{} on a quantized Transformer block. \texttt{attn.qkv}, \texttt{mlp.w12}
and \texttt{mlp.w3} apply a rotation matrix transform $R$ to their input, the matching inverse
folded offline into the weights. \texttt{attn.proj} needs no input-side transform, since its rotation
is absorbed entirely into $W_{\mathrm{proj}}$. $R$ is the group element addressing S3. The standard
token-uniform correction removes a shared systematic offset and can be folded into a layer bias.
The group-external term $b_{\mathrm{s}}$ targets the S1--S2 interaction by correcting only the special-token
row.}
\label{fig:framework}
\vspace{-16pt}
\end{figure*}

\section{Method}
\label{sec:method}

This section first fixes the host model and the shared-grid PTQ setting, then states four
pre-quantization properties. We next characterize the token-blind reparameterization group, decide
whether a row-selective term is needed, and assign rotation, token-uniform correction, and a
special-row term accordingly (Fig.~\ref{fig:framework}).

\subsection{Preliminaries}
\label{sec:pre}

\paragraph{MapAnything.}
Given $N$ RGB images $I_i$ and optional geometric inputs $g_i$, MapAnything predicts a \emph{factored}
representation $\{m,(R_i,\tilde{D}_i,\tilde{P}_i)_{i=1}^{N}\}$, where $R_i$ is the camera pose,
$\tilde{D}_i$ and $\tilde{P}_i$ are the up-to-scale depth and point map, and $m$ is a single global
metric scale. A frozen DINOv2 backbone~\citep{dinov2} converts each image into
$n$ patch tokens of width $d$, which are combined with encoded geometric cues by $\phi$. A learnable
\emph{scale token} $h_{\mathrm{s}}\in\mathbb{R}^{1\times d}$, the distinguished token in MapAnything, is then appended:
\begin{equation}
  H^{(0)}=[\phi(I_1,g_1);\ldots;\phi(I_N,g_N);h_{\mathrm{s}}],
  \qquad X_i^{\mathrm{metric}}=m\,\widetilde{X}_i ,
  \label{eq:mapanything}
\end{equation}
where $H^{(0)}$ is the initial token matrix and $\widetilde{X}_i$ denotes the up-to-scale geometry
$(R_i,\tilde{D}_i,\tilde{P}_i)$. $X_i^{\mathrm{metric}}$ is its metric counterpart.
A 16-layer alternating-attention Transformer interleaves \emph{frame} blocks, which act within a
single view, and \emph{global} blocks, which exchange information across views. Dense Prediction Transformer (DPT)~\citep{dpt} and
pose heads decode the patch tokens into $\widetilde{X}_i$, whereas a two-layer
multilayer perceptron (MLP) on the final scale token regresses $m$.

\paragraph{Post-training quantization.}
PTQ maps tensors to low-bit integers using a small calibration set $\calD_{\mathrm{calib}}$ and no
end-to-end fine-tuning. Quantizing a tensor $x$ at $b$ bits (step $\Delta$, bound $Q_b$) gives
\begin{equation}
  \widehat{x}=\Delta\,\operatorname{clamp}\!\left(
  \operatorname{round}(x/\Delta),-Q_b,Q_b\right),\qquad
  Q_b=2^{b-1}-1,\quad \Delta=\|x\|_\infty/Q_b .
  \label{eq:quantizer}
\end{equation}
We quantize the four linear maps of each block, the query--key--value (QKV) projection, the attention
output projection, and the two projections of the gated feed-forward network (FFN) (Fig.~\ref{fig:framework}), while the
multi-modal encoders, LayerNorm, softmax, the DPT head, pose heads, and scale MLP stay in floating
point, following common practice~\citep{smoothquant,quarot}. Weights use per-output-channel scales.
Activations use a \emph{per-tensor} static scale calibrated offline, so all token rows of a layer
share one activation grid. Following block-wise PTQ~\citep{adaround,brecq,qdrop,omniquant},
calibration minimizes
\begin{equation}
  \mathcal{L}_{\mathrm{quant}}
  =\mathbb{E}_{x\sim\calD_{\mathrm{calib}}}\bigl\|F(W,X)-F(\widehat{W},\widehat{X})\bigr\|_2^2,
  \label{eq:blockmse}
\end{equation}
over floating-point and quantized outputs of a module $F$ with weights $W$ and activations $X$.

\subsection{Four Structural Properties}
\label{sec:props}

We state each property for a generic Transformer processing tokens $H\in\mathbb{R}^{T\times d}$ of
sequence length $T$ and width $d$, with a distinguished row $h_{\mathrm{s}}$, and indicate the
measurement that certifies it. All such
measurements precede quantization, and the values obtained on the three models we study are collected
in Sec.~\ref{sec:exp-transfer}.

\begin{figure*}[t]
\centering
\begin{subfigure}[t]{0.32\textwidth}
\centering\includegraphics[width=\linewidth]{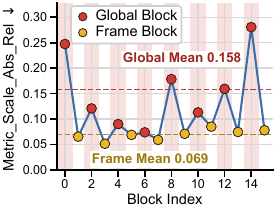}
\caption{}
\end{subfigure}\hfill
\begin{subfigure}[t]{0.32\textwidth}
\centering\includegraphics[width=\linewidth]{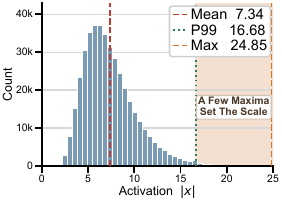}
\caption{}
\end{subfigure}\hfill
\begin{subfigure}[t]{0.32\textwidth}
\centering\includegraphics[width=\linewidth]{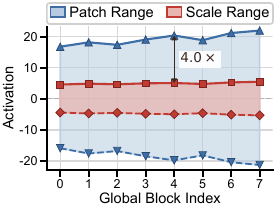}
\caption{}
\end{subfigure}
\vspace{-7pt}
\caption{Structural properties of MapAnything that guide the PTQ recipe.
(a): block-wise relative scale error. Global blocks (red) are more sensitive than frame blocks,
consistent with S4.
(b): histogram of flattened activation magnitudes $|x|$ in \texttt{attn.qkv}. A few maxima set the
per-tensor scale (S3).
(c): patch vs.\ scale activation ranges. The type-level gap supports the shared-grid mismatch in S1.
S2 follows from the multiplicative role of the scale token.}
\label{fig:diagnosis}
\vspace{-12pt}
\end{figure*}

\vspace{-3pt}
\paragraph{(S1) Shared grid across heterogeneous token types.}
One step $\Delta$ is shared by tokens whose dynamic ranges differ by \emph{type} rather than content.
With occupancy $\rho_t=\max_c|x_{t,c}|/\max_{t',c}|x_{t',c}|$, where $x_{t,c}$ is channel $c$ of
token $t$, token $t$ reaches only
$\rho_t(2^{b-1}\!-\!1)$ of the available levels, i.e.\ the effective bit-width
\begin{equation}
  b_{\mathrm{eff}}(t)=b-\log_2\bigl(1/\rho_t\bigr).
  \label{eq:beff}
\end{equation}
The property is certified by the type-level range ratio $\max_t\rho_t/\rho_{\mathrm{s}}$, which in
MapAnything grows by two orders of magnitude with depth and leaves the scale token a fraction of a
bit at deployment precision (Fig.~\ref{fig:diagnosis}(c)). Eq.~\eqref{eq:beff} is a descriptive
statistic of the grid. Sec.~\ref{sec:group} instead analyzes the limits of token-uniform
reparameterizations.

\vspace{-3pt}
\paragraph{(S2) Single token with multiplicative leverage.}
Although the distinguished token occupies only $1/T$ of the sequence, its perturbation propagates to the
\emph{whole} output. In MapAnything, $h_{\mathrm{s}}$ is the sole carrier of $m$. Holding
$\widetilde X_i$ fixed and isolating perturbations along the metric-scale pathway,
Eq.~\eqref{eq:mapanything} gives
$\|\hat{X}^{\mathrm{metric}}_i-X^{\mathrm{metric}}_i\|/\|X^{\mathrm{metric}}_i\|=|\hat m-m|/m$ for
every $i$, whereas a patch token's error is local and averaged by the dense head. Two distinctions
matter when auditing other models. First, S2 is \emph{multiplicative, output-wide} leverage, not mere
gradient concentration. A token consumed by a dedicated head carries most of the task gradient by
construction, in floating-point as much as in integer arithmetic, while its error stays confined to
one output branch. Second, S1 characterizes vulnerability, whereas S2 characterizes consequence. The
two are independent and Sec.~\ref{sec:gamma} requires both. Jointly they produce
\emph{differentiated degradation}, which single-metric evaluation cannot detect. S2 is also the
property up-to-scale predictors lack; hence, a recipe tuned on them need not transfer.

\vspace{-3pt}
\paragraph{(S3) Heavy-tailed channel energy.}
Activation energy concentrates in a few coordinates, so an axis-aligned per-tensor grid is set by rare
peaks. Fig.~\ref{fig:diagnosis}(b) histograms the flattened magnitudes $|x|$ at \texttt{attn.qkv}. Most
mass lies well below the 99th percentile, while a few maxima determine the shared absmax. S3 is what
rotation was designed for, and is shared by essentially all Transformers, so it cannot by itself
distinguish models.

\paragraph{(S4) Sparse routing of the distinguished token.}
The special token is written by a strict subset $\mathcal{B}_{\mathrm{g}}$ of blocks, certified by
$\|h^{(\ell+1)}_{\mathrm{s}}-h^{(\ell)}_{\mathrm{s}}\|>0$, where $\ell$ indexes the block. In MapAnything, the frame blocks pass the
scale token through unchanged, so only the global blocks update it, as in
Fig.~\ref{fig:diagnosis}(a). This halves the insertion points.

S3 acts along the channel axis of $H$, while S1--S2 act along the token axis. S4 then restricts
where that token-axis correction is applied.

\subsection{What the Reparameterization Group Can and Cannot Do}
\label{sec:group}

At a linear layer $y=xW+b$ with layer bias $b$ (not the bit-width), the transformations leaving the floating-point map invariant are
\begin{equation}
  x'=(x+a)M,\quad W'=M^{-1}W,\quad b'=b-aW,
  \qquad M\in\mathrm{GL}(d),\ a\in\mathbb{R}^{1\times d}.
  \label{eq:group}
\end{equation}
Every PTQ method based on rotation or smoothing is an element of Eq.~\eqref{eq:group}. Two facts drive
everything below. First, $M$ acts by right multiplication, so the same linear operator is applied to every token row. Second, $a$ is a single row vector broadcast across tokens, so every row receives the same offset. Hence, the group has no row-indexed degree of freedom.
To compare tokens under a shared step $\Delta$, we use the high-resolution quantization-noise proxy
$\mathrm{SNR}_{\Delta}(t)\triangleq\|h_t\|_2^2/(d\Delta^2/12)$, whose denominator is the standard
uniform-rounding model. The proposition concerns this proxy, not any token's realized error.

\begin{proposition}[Common-mode rescaling of the SNR proxy]
\label{prop:common}
Let $R$ be an orthogonal rotation and $\Delta,\Delta_R$ the shared steps before and after $H\mapsto HR$. Then
$\mathrm{SNR}_{\Delta_R}(t)=(\Delta/\Delta_R)^2\,\mathrm{SNR}_{\Delta}(t)$ for every $t$, and
$(\Delta/\Delta_R)^2\le d$. In particular, for all token indices $i,j$,
\begin{equation}
  \frac{\mathrm{SNR}_{\Delta_R}(i)}{\mathrm{SNR}_{\Delta_R}(j)}
  =\frac{\mathrm{SNR}_{\Delta}(i)}{\mathrm{SNR}_{\Delta}(j)}
  =\frac{\|h_i\|_2^2}{\|h_j\|_2^2}.
  \label{eq:invariant}
\end{equation}
\end{proposition}

\vspace{-4pt}
Reducing the shared step therefore gives every token the same multiplicative proxy gain while
leaving all pairwise ratios unchanged, so
$\kappa_{\mathrm{s}}\triangleq\|h_{\mathrm{s}}\|_2/\max_t\|h_t\|_2$ is a property of the learned
representation, not of the orthogonal basis. The proxy is not exact in the sub-step regime. When
$\rho_tQ_b<1$, a token occupies fewer than one level and may be partly or entirely rounded to zero.
We therefore measure realized residuals directly when estimating affine correctability below.

\begin{proposition}[Conditioning ceiling for general elements]
\label{prop:cond}
Let $M\in\mathrm{GL}(d)$ with condition number $\mathrm{cond}(M)=\sigma_{\max}/\sigma_{\min}$. Then for all $i,j$,
$\|h_iM\|_2/\|h_jM\|_2\in[\mathrm{cond}(M)^{-1},\mathrm{cond}(M)]\cdot\|h_i\|_2/\|h_j\|_2$, so $M$ can change any pairwise token-norm ratio by at most a factor of $\mathrm{cond}(M)$.
\end{proposition}

\vspace{-4pt}
An ill-conditioned $M$ also changes the dynamic range of $W'=M^{-1}W$ and can increase its
quantization error. We treat this as a practical trade-off rather than claiming that weight-side quantization error must increase at a matching rate, since
the realized error depends on the weight distribution and granularity.

\begin{lemma}[Absence of row-selective freedom]
\label{lem:row}
$\mathrm{cond}(M)=1$ recovers Prop.~\ref{prop:common} with zero token-ratio improvement. A general
$M$ can move that ratio within Prop.~\ref{prop:cond}'s ceiling, but every element of
Eq.~\eqref{eq:group} is row-uniform, so none can implement a correction on token identity.
The minimal \emph{additive} row-selective extension is
$u_{\mathrm{s}}b_{\mathrm{s}}^\top$, with $u_{\mathrm{s}}$ the special-row indicator.
\end{lemma}

Per-token activation scaling is a second row-dependent option outside Eq.~\eqref{eq:group}. It
changes the quantizer rather than reparameterizing the linear map, assigning token-specific steps
that improve grid occupancy. In contrast, $u_{\mathrm{s}}b_{\mathrm{s}}^\top$ leaves the grid
unchanged and removes the special row's systematic residual. The two address different aspects of
the error and entail different backend trade-offs.

\begin{proposition}[Exact value of an affine correction]
\label{prop:affine}
Let $e=\hat h-h$ be the quantization error at an insertion point, $e=\bar e+\varepsilon$ with
$\bar e=\mathbb{E}_{\calD_{\mathrm{calib}}}[e]$ and $\mathbb{E}[\varepsilon]=0$. Then
$\min_c\mathbb{E}\|e-c\|_2^2$ is attained at $c=\bar e$ with value $(1-\eta)\mathbb{E}\|e\|_2^2$,
$\eta=\|\bar e\|_2^2/\mathbb{E}\|e\|_2^2\in[0,1]$. A token-uniform correction removes only
$\|\mathbb{E}_t[\bar e_t]\|_2^2$, leaving the type-specific residual
$\bar e_{\mathrm{s}}-\mathbb{E}_t[\bar e_t]$ intact.
\end{proposition}

An affine correction therefore helps only to the extent that the error has a systematic component
$\bar e$, of which $\eta$ is the fraction. Proofs are in App.~\ref{app:proofs}.

\subsection{$\Gamma$: Diagnosing Need and Affine Correctability}
\label{sec:gamma}

The first-order loss perturbation from token $t$ is bounded by
$|\langle\partial\mathcal{L}/\partial h_t,e_t\rangle|\le\lambda_t\|e_t\|_2$. Here, $\mathcal{L}$ is
the task loss and $\lambda_t=\|\partial\mathcal{L}/\partial h_t\|_2$ measures the token's output leverage. Under the
high-resolution proxy, all tokens share the same quantization-noise denominator, so
$\lambda_t/\|h_t\|_2$ is a relative vulnerability score. Rotation reduces the common step but leaves
the norm-ratio component unchanged (Prop.~\ref{prop:common}). For a distinguished token that passes
the S2 audit, we therefore define
\begin{equation}
  \Gamma
  = \underbrace{\frac{\lambda_{\mathrm{s}}}{\bar{\lambda}_{\mathrm{patch}}}}_{\text{output criticality}}
  \cdot
  \underbrace{\Bigl(\tfrac{\max_t\|h_t\|_2}{\|h_{\mathrm{s}}\|_2}-1\Bigr)_{\!\!+}}_{\text{differential grid deficit}},
  \label{eq:gamma}
\end{equation}
where $(x)_+=\max(x,0)$ and $\bar{\lambda}_{\mathrm{patch}}$ is the mean patch leverage. $\Gamma$ needs
one forward and one backward pass per scene, without quantizing the model or running a benchmark. It is a
screening statistic, not a calibrated error bound. In the sub-step regime, the realized error can
depart from the proxy, so $\Gamma$ is used only to screen, and the affine remedy is checked on
calibration residuals.

Neither factor suffices alone. High leverage may merely reflect dedicated-head routing with no grid
disadvantage, while a small token is irrelevant if the output path does not read it
(Sec.~\ref{sec:exp-transfer}). $\Gamma\!\approx\!0$ thus predicts little need beyond the common-mode
treatment, whereas $\Gamma\!\gg\!0$ identifies a token that is both quantization-disadvantaged and
output-critical. It does \emph{not} by itself imply an affine correction suffices. After fitting the
token-uniform term $\beta$, let $r_{\mathrm{s}}=e_{\mathrm{s}}-\mathbb{E}_t[\bar e_t]$ be the leftover residual and
$\eta_{\mathrm{s}\mid\beta}=\|\mathbb{E}[r_{\mathrm{s}}]\|_2^2/\mathbb{E}\|r_{\mathrm{s}}\|_2^2$, the share recoverable by $b_{\mathrm{s}}$
(Prop.~\ref{prop:affine}).

\subsection{\ours{}: Structure-Guided Mechanisms}
\label{sec:mechanisms}

For S3, we take an orthogonal rotation $R$ as the group element in Eq.~\eqref{eq:group}. Among
standard constructions, App.~\ref{app:rot} shows that a randomized Hadamard transform (RHT) attains the
strongest common-mode gain, so we instantiate $R=D\mathcal{H}$ with head-sized
blocks as in \citet{quarot} and a full transform on the FFN hidden axis, where $\mathcal{H}$ is a
Hadamard matrix and $D$ a random sign diagonal. For a block of width $p$,
concentration gives $\|xD\mathcal{H}\|_\infty=O(\|x\|_2\sqrt{\log p/p})$ w.h.p. The realized
reduction in $\Delta$ additionally depends on the input's pre-rotation coherence and, for
block-diagonal transforms, its energy across blocks. $R^\top W$ is folded offline. The online
transform remains only at QKV and the two FFN projections.

The token-uniform term $\beta$ is standard bias correction. It removes the mean error shared across
tokens but changes neither the shared grid nor its type-level range mismatch. It can be folded into
an existing layer bias. Otherwise, it is stored as one output-width vector. The row-selective term
$b_{\mathrm{s}}$ addresses the S1--S2
interaction by removing the remaining systematic error on the distinguished row and lies outside the
reparameterization group (Lemma~\ref{lem:row}):
\begin{equation}
  \beta^{(\ell)\star}=-\,\mathbb{E}_{t}\bigl[\bar e^{(\ell)}_t\bigr],
  \qquad
  b_{\mathrm{s}}^{(\ell)\star}
  =\mathbb{E}_{\calD_{\mathrm{calib}}}\bigl[h^{(\ell)}_{\mathrm{s}}-\hat h^{(\ell)}_{\mathrm{s}}\bigr]
  -\beta^{(\ell)\star}.
  \label{eq:bias}
\end{equation}
It is omitted when $\Gamma\!\approx\!0$.
$\eta_{\mathrm{s}\mid\beta}$ quantifies whether this affine form provides material recovery. With $U$ the all-ones vector, the combined correction is
$B^{(\ell)}(H)=U(\beta^{(\ell)})^\top+u_{\mathrm{s}}(b_{\mathrm{s}}^{(\ell)})^\top$, applied as
$\widetilde H^{(\ell)}=\hat H^{(\ell)}+B^{(\ell)}(H)$.

Following S4, at width $m_\ell$ (not scale $m$), $b_{\mathrm{s}}^{(\ell)}$ is a single-row $O(m_\ell)$ addition that leaves
integer general matrix-matrix multiplication (GEMM) unchanged. We apply $\beta$ after all four linear maps and, by S4, $b_{\mathrm{s}}$ only in
$\mathcal{B}_{\mathrm{g}}$. The four output widths are $3d$, $d$, $2d_{\mathrm{ff}}$ and $d$, with $d_{\mathrm{ff}}$ the FFN width, so each
block holds $5d+2d_{\mathrm{ff}}$ correction scalars and the explicit row-selective storage is
$|\mathcal{B}_{\mathrm{g}}|(5d+2d_{\mathrm{ff}})$. Both terms are calibrated after rotation.

\section{Experiments}
\label{sec:exp}

\subsection{Experimental Setup}
\label{sec:exp-setup}

\textbf{Evaluation settings.}
We evaluate on ETH3D~\citep{eth3d}, ScanNet++v2 (SN++v2)~\citep{scannetpp}, and
TartanAirV2-WB (TAV2)~\citep{tartanair,ufm}, covering indoor, outdoor, and synthetic geometry.
Results are macro-averaged across the three datasets under $N\!=\!2$ and $N\!\in\!\{8,16\}$ with the official MapAnything
sampling~\citep{mapanything}. Main tables use images only. Because S2 predicts differentiated
degradation, we report a \emph{metric} family (scale rel.\ error, absolute trajectory error (ATE)) and an \emph{up-to-scale}
family (point and depth rel.\ error and $\tau$, area under the pose-accuracy curve (AUC@$30^\circ$), ray error), so a collapse of $m$
is not masked by scale-aligned scores. Bit-widths are W4A8 and W4A4. The component ablation and the checks on VGGT and DA3 use W4A4, where the gap between mechanisms is largest. We measure accuracy, latency,
and peak memory consumption on a single NVIDIA H20 GPU (96\,GB). Calibration uses 64 samples at
$518{\times}518$ from the MegaTrain mixture~\citep{mapanything} and contains no ETH3D data, so that set also probes
cross-dataset generalization. Further details are in App.~\ref{app:impl}.

\textbf{Baselines.}
We re-implement round-to-nearest (RTN), SmoothQuant~\citep{smoothquant}, and QuaRot~\citep{quarot} on MapAnything, using the same quantized layers and calibration data as \ours{}. By Sec.~\ref{sec:group} they are elements of
Eq.~\eqref{eq:group}, so the comparison tests the group ceiling rather than implementation quality.
Rotation and smoothing can recover common-mode error, but not a token-selective residual.

\Needspace{18\baselineskip}
\subsection{Camera Pose Estimation}
\label{sec:exp-pose}

\begin{wraptable}{r}{0.62\linewidth}
\vspace{-6pt}
\centering
\caption{Camera pose estimation, averaged over ETH3D, SN++v2 and TAV2.}
\label{tab:pose}
\small
\setlength{\tabcolsep}{4pt}
\begin{tabular}{@{}lcccc@{}}
\toprule
\multirow{2}{*}[-0.6ex]{Method} & \multirow{2}{*}[-0.6ex]{W/A}
  & frames=2 & frames=8 & frames=16 \\
\cmidrule(lr){3-3}\cmidrule(lr){4-4}\cmidrule(lr){5-5}
  & & AUC@30$^\circ$ & AUC@30$^\circ$ & AUC@30$^\circ$ \\
\midrule
Full Prec. & 16/16 & 98.9 & 99.2 & 98.8 \\
\midrule
RTN & 4/8 & 91.8 & 89.2 & 88.1 \\
SmoothQuant & 4/8 & 96.9 & 94.9 & 94.1 \\
QuaRot & 4/8 & 98.5 & 98.5 & 98.1 \\
\oursrow \textbf{\ours{}} & 4/8 & \textbf{98.6} & \textbf{98.6} & \textbf{98.1} \\
\midrule
RTN & 4/4 & 71.8 & 55.3 & 51.0 \\
SmoothQuant & 4/4 & 73.5 & 56.2 & 51.4 \\
QuaRot & 4/4 & 88.0 & 84.8 & 83.7 \\
\oursrow \textbf{\ours{}} & 4/4 & \textbf{92.8} & \textbf{87.4} & \textbf{86.5} \\
\bottomrule
\end{tabular}
\vspace{-8pt}
\end{wraptable}
We conduct camera pose estimation experiments on ETH3D, SN++v2, and TAV2 using 2, 8, and 16 input
views, and report AUC@30$^\circ$ macro-averaged over the three datasets. This metric jointly measures
relative rotation and translation accuracy. Notably, ETH3D is excluded from calibration and
therefore also evaluates cross-dataset generalization. Under the relatively moderate W4A8 setting,
\ours{} preserves camera pose accuracy with near-lossless performance. Under the more aggressive
W4A4 setting, its gap to full precision grows progressively with the number of input views. This
trend agrees with S1. As the view count increases, more patch tokens compete to determine the shared
absmax, further reducing the effective quantization resolution of the distinguished pathway. The
ordering among baselines further validates our mechanism analysis. RTN and SmoothQuant do not
adequately address the heavy-tailed channel distribution in S3 and therefore fail to recover the
dominant common-mode loss. QuaRot substantially reduces the overall quantization error through
orthogonal rotation, but its token-blind transformation cannot alter the relative SNR between
special and patch tokens. The additional improvement of \ours{} thus stems from explicitly
correcting differentiated errors along the token axis, rather than from dataset-specific tuning or
mismatched quantization settings.

\Needspace{18\baselineskip}
\subsection{Two-View Reconstruction: Testing for Differentiated Degradation}
\label{sec:exp-recon}

\begin{wraptable}{r}{0.66\linewidth}
\vspace{-10pt}
\centering
\caption{Two-view reconstruction, averaged over ETH3D, SN++v2 and TAV2.}
\label{tab:recon}
\small
\setlength{\tabcolsep}{2.2pt}
\begin{tabular}{@{}lccccccccc@{}}
\toprule
\multirow{2}{*}[-0.4ex]{Method} & \multirow{2}{*}[-0.4ex]{W/A}
  & \textbf{Scale} & \multicolumn{2}{c}{\textbf{Points}} & \multicolumn{2}{c}{\textbf{Pose}}
  & \multicolumn{2}{c}{\textbf{Depth}} & \textbf{Rays} \\
\cmidrule(lr){3-3}\cmidrule(lr){4-5}\cmidrule(lr){6-7}\cmidrule(lr){8-9}\cmidrule(lr){10-10}
  & & rel$\downarrow$ & rel$\downarrow$ & $\tau\uparrow$
  & ATE$\downarrow$ & AUC$\uparrow$ & rel$\downarrow$ & $\tau\uparrow$ & err$^\circ\downarrow$ \\
\midrule
Full Prec. & 16/16 & 0.05 & 0.06 & 60.3 & 0.01 & 98.9 & 0.06 & 57.3 & 0.19 \\
\midrule
RTN & 4/8 & 0.10 & 0.09 & 43.1 & 0.02 & 91.8 & 0.08 & 37.4 & 0.57 \\
SmoothQuant & 4/8 & 0.06 & 0.07 & 52.5 & 0.02 & 96.9 & 0.06 & 48.1 & 0.34 \\
QuaRot & 4/8 & 0.05 & 0.06 & 59.1 & 0.01 & 98.5 & 0.06 & 55.2 & 0.22 \\
\oursrow \textbf{\ours{}} & 4/8 & \textbf{0.05} & \textbf{0.06} & \textbf{59.2} & \textbf{0.01} & \textbf{98.6} & \textbf{0.06} & \textbf{56.0} & \textbf{0.19} \\
\midrule
RTN & 4/4 & 0.48 & 0.30 & 13.0 & 0.11 & 73.3 & 0.29 & 9.2 & 5.86 \\
SmoothQuant & 4/4 & 0.35 & 0.32 & 10.1 & 0.09 & 73.5 & 0.29 & 8.7 & 3.05 \\
QuaRot & 4/4 & 0.20 & 0.18 & 33.2 & 0.05 & 83.1 & 0.15 & 34.5 & 0.71 \\
\oursrow \textbf{\ours{}} & 4/4 & \textbf{0.12} & \textbf{0.09} & \textbf{43.4} & \textbf{0.02} & \textbf{92.7} & \textbf{0.07} & \textbf{41.0} & \textbf{0.45} \\
\bottomrule
\end{tabular}
\vspace{-8pt}
\end{wraptable}
We evaluate two-view reconstruction to test whether the quantized model preserves
the full geometric output. Scale relative error and ATE measure the metric path,
while point maps, pose, depth, and rays are scored after scale alignment.
Lower relative error, ATE, and ray error indicate better quality. Higher inlier
ratio $\tau$ and pose AUC indicate more accurate reconstruction.
Under W4A8, \ours{} attains the best quantized results on every metric and stays
near full precision on pose, dense geometry, and rays. Under W4A4, metric errors grow faster than
up-to-scale scores, as S2 predicts. Relative to QuaRot, \ours{} still reduces scale error from
$0.20$ to $0.12$ and raises point inliers from $33.2$ to $43.4$.

\FloatBarrier
\Needspace{16\baselineskip}
\subsection{Ablation: Removing One Mechanism at a Time}
\label{sec:exp-ablation}

\begin{wraptable}{r}{0.48\linewidth}
\vspace{-6pt}
\centering
\caption{Component ablation under W4A4. 
\textbf{R}: rotation, $\boldsymbol{\beta}$: token-uniform bias correction,
$\boldsymbol{b_{\mathrm{s}}}$: row-selective correction.}
\label{tab:ablation}
\small
\setlength{\tabcolsep}{3pt}
\renewcommand{\arraystretch}{1.08}
\begin{tabular}{@{}cccrrr@{}}
\toprule
\multirow{2}{*}[-0.4ex]{R} & \multirow{2}{*}[-0.4ex]{$\beta$} & \multirow{2}{*}[-0.4ex]{$b_{\mathrm{s}}$}
  & ETH3D & SN++v2 & TAV2 \\
 & & & AUC@30$\uparrow$ & AUC@30$\uparrow$ & AUC@30$\uparrow$ \\
\midrule
$\times$ & $\times$ & $\times$ & 77.0 & 74.3 & 64.0 \\
\checkmark & $\times$ & $\times$ & 83.7 & 86.9 & 86.1 \\
\checkmark & $\times$ & \checkmark & 87.3 & 88.5 & 89.1 \\
\checkmark & \checkmark & $\times$ & 88.2 & 90.4 & 89.4 \\
\oursrow \checkmark & \checkmark & \checkmark & \textbf{91.4} & \textbf{93.7} & \textbf{93.2} \\
\bottomrule
\end{tabular}
\vspace{-8pt}
\end{wraptable}
\textbf{Mechanism ablation.}
We conduct W4A4 two-view ablations to isolate the three components of \ours{}
(Table~\ref{tab:ablation}). Rotation provides the largest first-stage recovery over naive PTQ,
consistent with its treatment of the heavy-tailed channel distribution in S3. With $R$ fixed, either
the standard token-uniform correction $\beta$ or the proposed row-selective correction $b_{\mathrm{s}}$
improves all three datasets, showing that they remove distinct residual components. Adding $b_{\mathrm{s}}$ on
top of $R+\beta$ yields a further $3.2$--$3.8$ AUC@30$^\circ$ and produces the best result on every
benchmark. The full configuration reaches $91.4$, $93.7$, and $93.2$, confirming that common bias
correction does not subsume the special-row residual targeted by $b_{\mathrm{s}}$.

\FloatBarrier
\subsection{Efficiency Analysis}
\label{sec:exp-eff}

Efficiency is evaluated under W8A8 because TensorRT provides fused INT8 kernels. W4 configurations
use fake quantization for accuracy evaluation.

\begin{figure}[!t]
\vspace{-4pt}
\begin{minipage}[t]{0.56\textwidth}
\vspace{0pt}
\centering
\includegraphics[width=\linewidth]{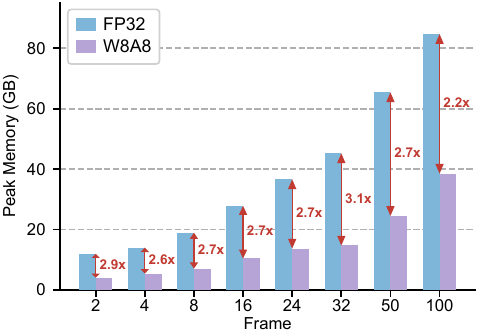}
\vspace{-22pt}
\captionof{figure}{Peak GPU memory versus the number of input frames on TAV2, comparing FP32 with W8A8.}
\label{fig:memory}
\end{minipage}\hfill
\begin{minipage}[t]{0.41\textwidth}
\vspace{0pt}
\captionof{table}{End-to-end latency (relative to the FP32 baseline) and pose AUC@30 on two-view
TAV2 at $518{\times}518$ image resolution. Automatic mixed precision (AMP) uses bfloat16 (BF16), while
the quantized rows use W8A8.}
\label{tab:latency}
\centering
\small
\setlength{\tabcolsep}{3pt}
\renewcommand{\arraystretch}{1.15}
\begin{tabularx}{\linewidth}{@{}lYYY@{}}
\toprule
Method & W/A & Latency $\downarrow$ & AUC@30 $\uparrow$ \\
\midrule
Baseline & 32/32 & $1.00\times$ & 98.9 \\
AMP & 16/16 & $0.52\times$ & 98.7 \\
Naive & 8/8 & $0.32\times$ & 91.9 \\
\oursrow \textbf{Ours} & 8/8 & $0.32\times$ & \textbf{98.9} \\
\bottomrule
\end{tabularx}
\end{minipage}
\vspace{2pt}

\begin{minipage}{\textwidth}
\captionof{table}{Audit columns are measured on FP checkpoints before quantization. Accuracy columns
use W4A4, two-view ETH3D pose AUC@$30^\circ$ $\uparrow$.}
\label{tab:transfer}
\centering
\small
\setlength{\tabcolsep}{2.2pt}
\begin{tabular}{@{}lccccccc ccccc@{}}
\toprule
\multirow{2}{*}[-0.4ex]{Model} & \multicolumn{4}{c}{Structural audit} & \multicolumn{3}{c}{Criterion}
  & \multicolumn{5}{c}{W4A4 pose AUC@$30^\circ$} \\
\cmidrule(lr){2-5}\cmidrule(lr){6-8}\cmidrule(lr){9-13}
 & (S1) & (S2) & (S3) & (S4)
 & $1/\kappa_{\mathrm{s}}$ & $\lambda_{\mathrm{s}}/\bar\lambda_{\mathrm{patch}}$ & $\Gamma$
 & FP & naive & $+R$ & $+R{+}\beta$ & $+R{+}\beta{+}b_{\mathrm{s}}$ \\
\midrule
VGGT & \checkmark & $\times$ & \checkmark & $\times$
  & $0.94$ & $8.3{\times}10^{3}$ & $0$
  & $88.36$ & $68.42$ & $75.9$ & $79.01$ & $78.99$ \\
DA3 & \checkmark & $\times$ & \checkmark & $\times$
  & $2.68$ & $0$ & $0$
  & $90.33$ & $71.1$ & $77.12$ & $81.21$ & $81.25$ \\
MapAnything & \checkmark & \checkmark & \checkmark & \checkmark
  & $6.02$ & $2.69{\times}10^{4}$ & $\mathbf{1.35{\times}10^{5}}$
  & $98.9$ & $77.0$ & $83.7$ & $88.2$ & $\mathbf{91.4}$ \\
\bottomrule
\end{tabular}
\end{minipage}
\vspace{-12pt}
\end{figure}

\textbf{Peak memory.}
Relative to FP32, W8A8 cuts peak memory by up to $3.1\times$ and by $2.2\times$ at $100$ frames
(Fig.~\ref{fig:memory}). The ratio contracts because floating-point components, normalization,
softmax, prediction heads, and intermediate buffers grow with $Nn+1$ and occupy an increasing
share of the footprint, while the absolute saving still grows with the number of frames. At $100$
frames, FP32 occupies $88\%$ of the H20, whereas W8A8 leaves substantial head-room.

\textbf{Latency.}
Under fused W8A8 kernels, \ours{} matches naive PTQ at $0.32\times$ FP32 latency while restoring
full-precision AUC (Table~\ref{tab:latency}). These results show that low-bit quantization eases
memory-bandwidth pressure while improving inference efficiency.

\FloatBarrier
\subsection{Generality: Does the Audit Predict What Each Model Needs?}
\label{sec:exp-transfer}

To test generality, we further audit VGGT~\citep{vggt}
and DA3~\citep{da3}, neither of which informed our method design, against S1--S4 and compute
$\Gamma$ on their full-precision checkpoints. All three models share the same signed quantizer,
weight and activation granularity, rotation construction, calibration data source, and W4A4 evaluation
setting. MapAnything additionally uses a staged 40-step refinement, whereas
VGGT and DA3 use closed-form correction without gradient refinement. No accuracy-driven tuning is
performed on either transfer model. Although the audit prescribes omitting $b_{\mathrm{s}}$
on VGGT and DA3, we deliberately instantiate the full $B$ as a conservative falsification test.
All three models satisfy S1 and S3. They use per-tensor activation scales, and their within-token
$\max/99$th-percentile ($\mathrm{P99}$) ratios at the final residual are substantially above the reference value for a
Gaussian of the same width. Rotation is therefore expected to benefit all three. Both transfer
models fail S2 and S4. VGGT is an up-to-scale predictor with no multiplicative scalar controlling
the global output. Its camera token is read by a dedicated head from the first layer of every block,
so it is neither multiplicatively load-bearing nor sparsely routed. DA3 introduces a camera token,
but its intermediate-layer interface removes that token before the dense head. The evaluated pose
pathway instead regresses from dense rays, leaving the token with no effective output leverage.

VGGT's camera token is \emph{output-critical but not quantization-disadvantaged}. Although its
dedicated head concentrates the pose gradient on that token, $1/\kappa_{\mathrm{s}}\!=\!0.94$ shows
that its norm matches the strongest patch token, so the shared grid does not disadvantage it. DA3
is the mirror image. Its camera token is disadvantaged but not critical
($1/\kappa_{\mathrm{s}}\!=\!2.68$, $\lambda_{\mathrm{s}}\!=\!0$). Consequently, both yield
$\Gamma\!=\!0$. This distinction shows that gradient concentration induced by dedicated-head
routing exists in both floating-point and quantized models and is not itself a quantization
pathology.

Rotation restores VGGT and DA3 to $85.9\%$ and $85.4\%$ of full-precision AUC@30, as S3 predicts.
Standard correction $\beta$ adds $3.11$ and $4.09$ points, whereas $b_{\mathrm{s}}$ is negligible
($-0.02$ and $+0.04$), versus $+3.2$ on MapAnything. Thus, $\Gamma$ predicts the materiality of row-selective
correction rather than generic bias correction. The models separate the two axes. VGGT has high gradient leverage but no grid disadvantage, DA3 has a grid disadvantage but no effective output leverage, and MapAnything exhibits both, supporting S1--S4
as family-level properties.

\vspace{-2pt}
\section{Conclusion}
\label{sec:concl}
\vspace{-3pt}

We identify four structural properties that determine when rotation-based PTQ needs token-selective
treatment beyond common-mode rotation, and prove that FP-preserving reparameterizations lack
row-indexed freedom. Guided by this analysis, we propose $\Gamma$ to screen this need and
$\eta_{\mathrm{s}\mid\beta}$ to test affine correctability. The resulting framework, \ours{}, achieves
near-lossless W4A8 quantization with a substantially reduced memory footprint. The row-selective
term helps high-$\Gamma$ models and is negligible when $\Gamma\approx 0$.

\subsection*{AI use statement}
In this work, we used generative AI tools for drafting and editing manuscript text, improving clarity
of exposition, and assisting with \LaTeX{} formatting. We have not used generative AI tools to
generate synthetic experimental data, fabricate results, or produce unverified mathematical claims.
Additionally, we used generative AI tools for literature search suggestions and for drafting
boilerplate code related to plotting and table formatting. We have reviewed all AI-assisted text and
code. All experimental results, claims, and figures were verified by the authors. We take
responsibility for the final content of this work.

\subsection*{Ethics statement}
This work studies post-training quantization for feed-forward 3D reconstruction and does not involve
human subjects or personally identifiable information. Publicly available geometric benchmarks are
used for evaluation. We do not foresee direct harmful applications beyond those already associated
with generic 3D reconstruction systems. All authors adhere to the ICLR Code of Ethics.

\subsection*{Reproducibility statement}
Implementation and evaluation details are in Sec.~\ref{sec:exp} and App.~\ref{app:impl}. The
measurement protocol for $\Gamma$ and the structural audit is in App.~\ref{app:audit}. Proofs and
rotation controls are in Apps.~\ref{app:proofs},~\ref{app:rot}. Upon acceptance, we will
release code and configuration files to reproduce the reported pipeline and main tables.

\bibliography{iclr2027_conference}
\bibliographystyle{iclr2027_conference}

\clearpage
\appendix

\section{Proofs}
\label{app:proofs}

\paragraph{Proof of Prop.~\ref{prop:common}.}
Orthogonality gives $\|h_tR\|_2=\|h_t\|_2$ for every $t$. Substitution into the definition of the
high-resolution proxy gives
$\mathrm{SNR}_{\Delta_R}(t)/\mathrm{SNR}_{\Delta}(t)=(\Delta/\Delta_R)^2$ for every $t$.
Eq.~\eqref{eq:invariant} follows by division. For the bound, $\|x\|_\infty\ge\|x\|_2/\sqrt d$ gives
$\Delta_R\ge\max_t\|h_t\|_2/(\sqrt d(2^{b-1}\!-\!1))$ while
$\Delta\le\max_t\|h_t\|_2/(2^{b-1}\!-\!1)$, so $(\Delta/\Delta_R)^2\le d$. The lower bound for
randomized Hadamard used in Sec.~\ref{sec:mechanisms} is the width-$p$ concentration bound
$\|xD\mathcal{H}\|_\infty=O(\|x\|_2\sqrt{\log p/p})$ w.h.p.~\citep{quipsharp}. Converting it into a
step-size gain additionally depends on the pre-rotation $\ell_\infty/\ell_2$ ratio and, for
block-diagonal transforms, the distribution of energy across blocks. \qed

\paragraph{Proof of Prop.~\ref{prop:cond}.}
For invertible $M$ and any $h$, $\sigma_{\min}(M)\|h\|_2\le\|hM\|_2\le\sigma_{\max}(M)\|h\|_2$.
Applying these bounds separately to the numerator and denominator gives
$\mathrm{cond}(M)^{-1}\|h_i\|_2/\|h_j\|_2
\le\|h_iM\|_2/\|h_jM\|_2
\le\mathrm{cond}(M)\|h_i\|_2/\|h_j\|_2$. \qed

\paragraph{Proof of Lemma~\ref{lem:row}.}
If $\mathrm{cond}(M)=1$, then $M$ is a scalar multiple of an orthogonal matrix and the bound of
Prop.~\ref{prop:cond} collapses to equality, i.e.\ zero improvement. The additive part maps
$H\mapsto H+Ua^\top$, which adds the same vector to every row and therefore commutes with
any relabelling of the rows. Right multiplication by $M$ also commutes with row relabelling.
Therefore their composition has no parameter conditioned on row identity, even though a
non-orthogonal $M$ may change the norms of differently oriented rows. Any prescribed row-dependent
correction must have the form $H+\sum_t u_tc_t^\top$ with at least two distinct $c_t$, the minimal
instance being a single indicator row. \qed

\paragraph{Proof of Prop.~\ref{prop:affine}.}
$\mathbb{E}\|e-c\|_2^2=\mathbb{E}\|e\|_2^2-2\langle c,\bar e\rangle+\|c\|_2^2$ is strictly convex in
$c$ with minimizer $c=\bar e$ and minimum
$\mathbb{E}\|e\|_2^2-\|\bar e\|_2^2=(1-\eta)\mathbb{E}\|e\|_2^2$. If $c$ is constrained to be
identical across tokens, the same computation with $\bar e$ replaced by $\mathbb{E}_t[\bar e_t]$
gives minimum $\mathbb{E}\|e\|_2^2-\|\mathbb{E}_t[\bar e_t]\|_2^2$, so the unremoved part on row $\mathrm{s}$
is $\bar e_{\mathrm{s}}-\mathbb{E}_t[\bar e_t]$. \qed

\section{Measuring the Structural Audit and $\Gamma$}
\label{app:audit}

All audit quantities in Table~\ref{tab:transfer} are measured on FP checkpoints. (S1) is read off the
deployment configuration (per-tensor activation scaling) and certified by the type-level range ratio
$\max_t\rho_t/\rho_{\mathrm{s}}$ at each linear layer. (S2) requires a token whose perturbation
propagates multiplicatively to the entire output. We certify it from the model's output
factorization, not from gradient magnitude, and then quantify its consequence by the leverage ratio
$\lambda_{\mathrm{s}}/\bar\lambda_{\mathrm{patch}}$ with
$\lambda_t=\|\partial\mathcal{L}/\partial h_t\|_2$ accumulated over calibration scenes and evaluated
on the same output path used at test time (for DA3 this is the dense-ray pose path, on which the
camera token is not read). (S3) is certified by the per-token $\max/\mathrm{P99}$ channel ratio,
(S4) by checking which blocks change the distinguished token, i.e.\
$\|h^{(\ell+1)}_{\mathrm{s}}-h^{(\ell)}_{\mathrm{s}}\|>0$. Finally
$\kappa_{\mathrm{s}}=\|h_{\mathrm{s}}\|_2/\max_t\|h_t\|_2$ is averaged over insertion points and
scenes. For models with one distinguished token per view, we take $h_{\mathrm{s}}$ to be that token.
Total cost is one forward and one backward pass per scene over a handful of scenes. No quantization
is performed.

\section{Classical Rotations as Controls}
\label{app:rot}

For a data-independent, non-learned rotation, we isolate three factors. \textbf{(A) Orthogonality}.
Exact cancellation by $R^\top$, so the substitution is lossless in FP. \textbf{(B) Block-wise
flatness}. $|R_{ij}|=1/\sqrt p$ for all non-zero entries of a block of size $p$. \textbf{(C)
Random de-alignment}. Random signs break alignment between a fixed basis and the model's
persistent outlier channels. Randomness is not cosmetic. For any \emph{fixed} orthogonal $R$ there
exists an input aligned with a row of $R^\top$ whose image is one-hot, so (C) is necessary for a
distribution-free guarantee. All five rows of Table~\ref{tab:rot-factors} are elements of
Eq.~\eqref{eq:group} with $\mathrm{cond}(M)=1$, so by Lemma~\ref{lem:row} they differ only in how much
of the common-mode gain they realize, never in the token-norm ratio they leave behind.

\begin{table}[h]
\centering
\caption{Classical rotations as controls for factors (A)--(C).}
\label{tab:rot-factors}
\small
\begin{tabular}{lccc}
\toprule
Rotation & (A) Orth. & (B) Flat & (C) Rand. \\
\midrule
Identity                                   & $\times$ & $\times$ & $\times$ \\
Discrete cosine transform (DCT)            & \checkmark & $\times$ & $\times$ \\
Fixed Hadamard ($\mathcal{H}$)             & \checkmark & \checkmark & $\times$ \\
Haar random orthogonal                     & \checkmark & $\times$ & \checkmark \\
Randomized Hadamard ($D\mathcal{H}$)       & \checkmark & \checkmark & \checkmark \\
\bottomrule
\end{tabular}
\end{table}

\begin{figure}[h]
\centering
\includegraphics[width=0.72\linewidth]{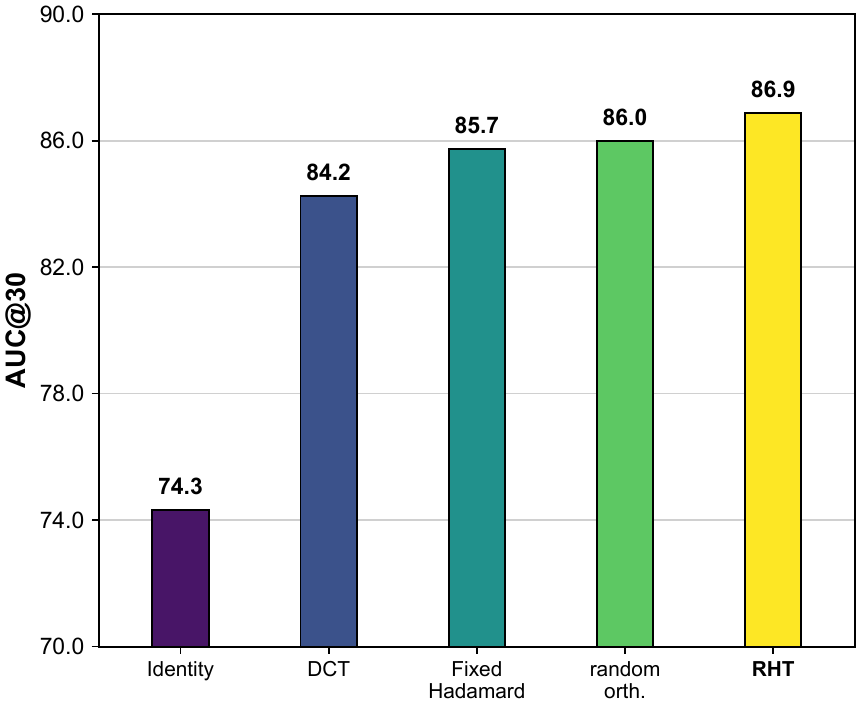}
\caption{How far the group ceiling can be pushed: transform ablation on SN++v2 (W4A4, bias
disabled).}
\label{fig:rot-ablation}
\end{figure}

\textbf{Locating the group ceiling.}
Prop.~\ref{prop:common} says the achievable common-mode gain depends on how well the transform
flattens $\|\cdot\|_\infty$. Fig.~\ref{fig:rot-ablation} measures how much of it each classical
choice realizes, with the quantizer and calibration set fixed and the bias disabled, so the endpoints
reproduce the naive and rotation-only rows of Table~\ref{tab:ablation} ($74.3$ and $86.9$). Orthogonality alone
accounts for most of the gain (DCT, $74.3\!\rightarrow\!84.2$), confirming that the axis-aligned
basis rather than the magnitude of the activations is what costs bits. Exact flatness adds $1.5$
points (fixed Hadamard, $85.7$) and random de-alignment $1.8$ (Haar, $86.0$), each worth little alone
since a flat but fixed basis can still align with persistent outlier channels. Only RHT provides all
three and is best ($86.9$), consistent with the concentration argument in
Sec.~\ref{sec:mechanisms}. The important reading is the gap
that remains. $86.9$ is where the \emph{best} group element saturates, and the $+6.8$ points the
combined correction $B$ then adds show that additive correction remains complementary to maximal
rotation.

\section{Implementation Details}
\label{app:impl}

\paragraph{Software and backend.}
Our accuracy environment uses Python 3.10.12, PyTorch 2.6.0+cu126, CUDA 12.6
(nvcc 12.6.85), cuDNN 9.5.1, NVIDIA driver 550.127.05, and
\texttt{horizon\_plugin\_pytorch} 3.1.3+cu126.torch260. Accuracy tables use PyTorch fake
quantization with outer AMP disabled. Runtime and peak memory measurements use TensorRT
10.8.0.43 on one NVIDIA H20 (96\,GB). Its internal INT GEMMs provide the low-bit kernels.
No standalone CUTLASS or \texttt{fast\_hadamard\_transform} package is used.

\paragraph{Quantization and rotation.}
We use symmetric signed absmax quantization with
$q_{\max}=2^{b-1}-1$ (INT4: $[-7,7]$, INT8: $[-127,127]$). Weight scales are static
per-output-channel and activation scales are static per-tensor. Main accuracy tables use W4A8 and W4A4.
The quantized scope comprises the four linear maps in each Transformer block. There are 64 modules for
MapAnything, 192 for VGGT, and 160 for DA3. We instantiate the rotation as an independent per-layer Walsh--Hadamard matrix preceded by a random sign diagonal $D$, with seed 0. The signs are applied first. It is folded into weights offline. Accuracy evaluation applies the remaining
online $xD\mathcal{H}$ in PyTorch, while TensorRT executes the exported efficiency model.

\paragraph{Calibration and bias refinement.}
Calibration uses 64 two-view samples at $518{\times}518$ from
\texttt{megatrain\_6d\_518\_many\_ar\_36ipg\_64g}, comprising
BlendedMVS~\citep{blendedmvs}, MPSD~\citep{mpsd}, ScanNet++v2~\citep{scannetpp},
Spring~\citep{spring}, TAV2-WB~\citep{tartanair,ufm}, and
UnrealStereo4K~\citep{unrealstereo4k} training data, with no ETH3D scenes.
The closed-form biases use one forward pass with damping 1. For MapAnything W4A4, we further
optimize each stage with Adam (learning rate $10^{-4}$, no weight decay, gradient clipping at
1.0) for 40 steps over three stages (blocks 0--7, 8--11, and 12--15). On VGGT and DA3, both terms of
$B$ are estimated in closed form over three rounds of 32 samples from the same loader, without
gradient refinement.

\paragraph{Held-out systematic residual diagnostic.}
For W4A4 with rotation, we split the 64 calibration scenes in half, fit $(\beta,b_{\mathrm{s}})$ on one half,
evaluate on the other, and swap the halves. At each of 32 global insertion points (eight blocks,
four linear maps), let $\mu_\beta=\mathbb{E}_t[\bar e_t]=-\beta^\star$ and
$r_{\mathrm{s}}=e_{\mathrm{s}}-\mu_\beta$. The conditional systematic fraction
$\eta_{\mathrm{s}\mid\beta}=\|\mathbb{E}[r_{\mathrm{s}}]\|_2^2/\mathbb{E}\|r_{\mathrm{s}}\|_2^2$ has median $0.311$
(interquartile range (IQR) $0.187$). Its energy-weighted value across insertion points is $0.387$. Applying the bias fit
on the opposite fold reduces held-out scale-token mean squared error (MSE) by $34.2\%$, yet directly measured whole-block
MSE by only $0.010\%$. Thus, a substantial predictable residual on the output-critical row is nearly
invisible to a token-averaged reconstruction loss.

\section{Additional Visual Comparisons}
\label{app:visual}

\begin{figure}[h]
\centering
\includegraphics[width=\linewidth]{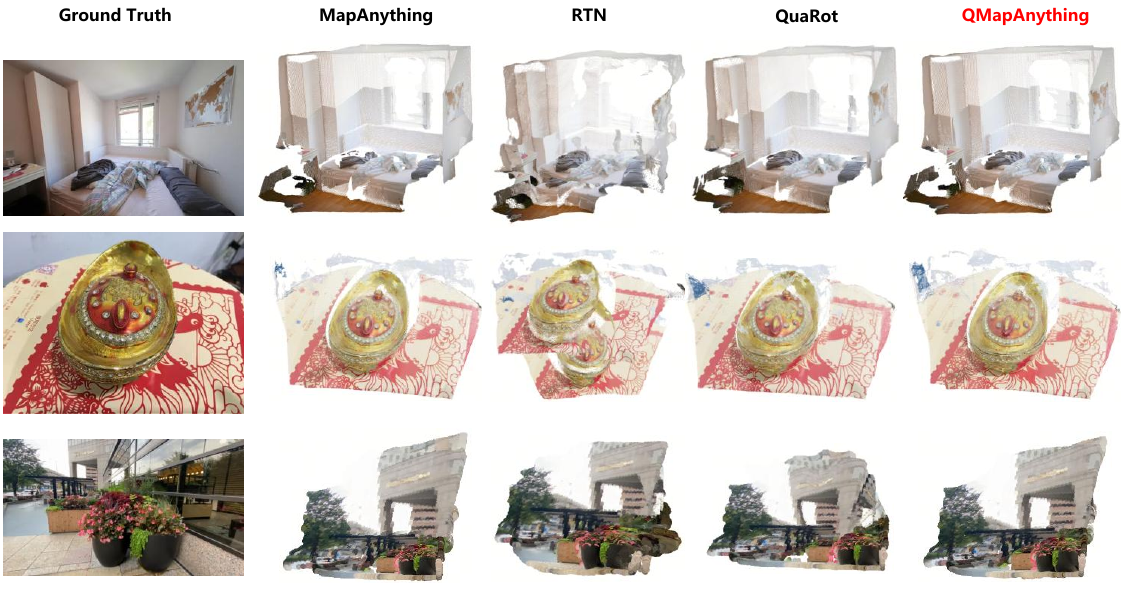}
\caption{Qualitative comparison on three scenes. MapAnything is the FP32 reconstruction. RTN, QuaRot,
and \ours{} are W4A4.}
\label{fig:visual}
\end{figure}

Fig.~\ref{fig:visual} compares reconstructions on three scenes. The first column is the input image.
MapAnything is the FP32 reference. RTN, QuaRot, and \ours{} use W4A4, the setting where the gap
between mechanisms is largest.

\end{document}